\documentclass[10pt,twocolumn,letterpaper]{article}

\usepackage[pagenumbers]{cvpr}  

\usepackage{capt-of}
\usepackage{xspace}
\usepackage{xcolor}
\usepackage{enumitem}
\usepackage{amsmath,amsthm,amssymb,amsfonts,dsfont,pifont,bm,bbm,mathrsfs,mathtools,nicefrac,extarrows,relsize}
\usepackage{algorithm,algpseudocode,listings}
\usepackage{booktabs,multirow,adjustbox,diagbox,threeparttable,tabularray,setspace}

\definecolor{cvprblue}{rgb}{0.21,0.49,0.74}
\usepackage[pagebackref,breaklinks,colorlinks,citecolor=cvprblue,bookmarks=false]{hyperref}
\usepackage{wrapfig}
\usepackage[capitalize]{cleveref}  

\crefname{section}{Sec.}{Secs.}
\Crefname{section}{Section}{Sections}
\crefname{appendix}{App.}{Apps.}
\Crefname{appendix}{Appendix}{Appendices}
\crefname{table}{Tab.}{Tabs.}
\Crefname{table}{Table}{Tables}
\crefname{figure}{Fig.}{Figs.}
\Crefname{figure}{Figure}{Figures}
\crefname{equation}{Eq.}{Eqs.}
\Crefname{equation}{Equation}{Equations}
\crefname{theorem}{Thm.}{Thms.}
\Crefname{theorem}{Theorem}{Theorems}
\crefname{lemma}{Lem.}{Lems.}
\Crefname{lemma}{Lemma}{Lemmas}
\crefname{remark}{Rem.}{Rems.}
\Crefname{remark}{Remark}{Remarks}
\crefname{corollary}{Cor.}{Cors.}
\Crefname{corollary}{Corollary}{Corollaries}
\crefname{algorithm}{Alg.}{Algs.}
\Crefname{algorithm}{Algorithm}{Algorithms}
\definecolor{lightred}{RGB}{200, 50, 50}   
\definecolor{lightblue}{RGB}{50, 100, 200} 
\definecolor{cellred}{RGB}{213, 123, 101}
\definecolor{cellgreen}{RGB}{0, 205, 0}
\definecolor{cellblue}{RGB}{54, 125, 189}
\definecolor{codegreen}{rgb}{0,0.6,0}
\definecolor{codegray}{rgb}{0.5,0.5,0.5}
\definecolor{codepurple}{rgb}{0.58,0,0.82}
\definecolor{backcolour}{rgb}{1.0,1.0,1.0}
\lstdefinestyle{mystyle}{
    backgroundcolor=\color{backcolour},
    commentstyle=\color{codegreen},
    keywordstyle=\color{magenta},
    numberstyle=\tiny\color{codegray},
    stringstyle=\color{codepurple},
    basicstyle=\ttfamily\scriptsize,
    breakatwhitespace=false,
    breaklines=true,
    captionpos=b,
    keepspaces=true,
    numbers=left,
    numbersep=5pt,
    showspaces=false,
    showstringspaces=false,
    showtabs=false,
    tabsize=2
}
\newcommand{\tocite}[1]{{\color{red} [TO CITE]}}

\newcommand{\methodname}{AvatarArtist}
\newcommand{\method}{\texttt{\methodname}\xspace}

\def\paperID{4979}
\def\confName{CVPR}
\def\confYear{2025}

\title{\methodname: Open-Domain 4D Avatarization  }

\author{
Hongyu Liu$^{1,2,*}$ \qquad
Xuan Wang$^{2, \S}$ \qquad
Ziyu Wan$^{3}$ \qquad
Yue Ma$^{1}$ \qquad
Jingye Chen$^{1}$ \qquad
Yanbo Fan$^{2}$ \\[3pt]
Yujun Shen$^{2}$ \qquad 
Yibing Song \qquad
Qifeng Chen$^{1,\S}$ \\[8pt]
$^{1}$HKUST \quad 
$^{2}$Ant Group \quad
$^{3}$City University of Hong Kong \\[5pt]
\url{https://kumapowerliu.github.io/AvatarArtist}
}

\newcommand\nonumfootnote[1]{%
\begingroup%
    \renewcommand\thefootnote{}\footnote{\hspace{-4pt}#1}%
    \addtocounter{footnote}{-1}%
\endgroup%
}

\begin{document}

\twocolumn[{
\renewcommand\twocolumn[1][]{#1}
\maketitle
\begin{center}
    \vspace{-5pt}
    \includegraphics[width=1.0\linewidth]{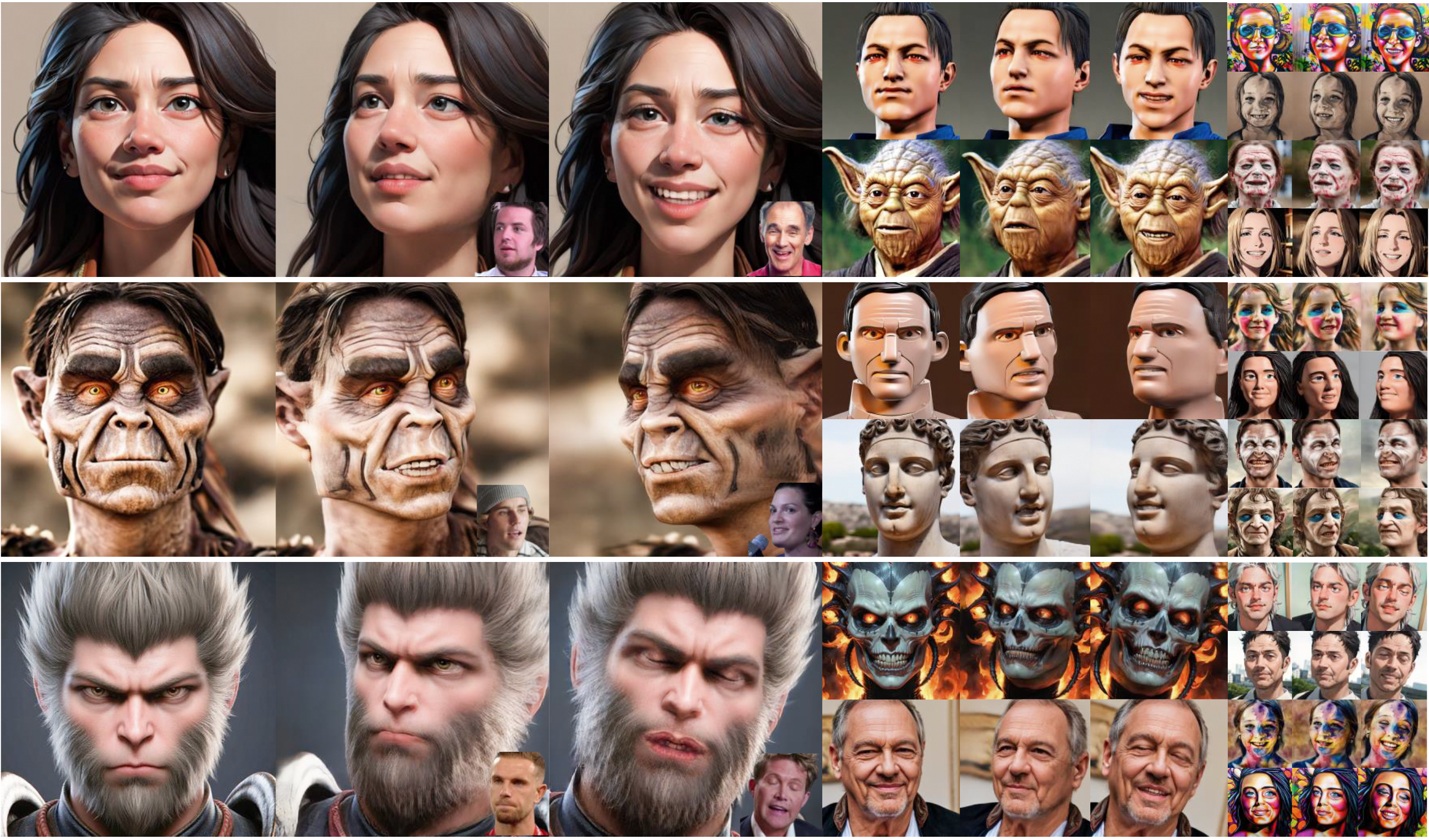}
    \vspace{-15pt}
    \captionsetup{type=figure}
    \caption{\textbf{Gallery of the proposed AvatarArtist}. Each row features several triplets, where the first column of each triplet is the source image. The subsequent two images in each triplet are results that follow the pose and expression of the driving image, as demonstrated in the bottom right corner of the first three columns. Specifically, our method is applicable to an open domain, encompassing a diverse range of categories including 3D cartoons, video game characters, sculptures, skulls, etc.
    }
    \label{fig:teaser}
    \vspace{15pt}
\end{center}
}]

\maketitle
\begin{abstract}

This work focuses on open-domain 4D avatarization, with the purpose of creating a 4D avatar from a portrait image in an arbitrary style.
We select parametric triplanes as the intermediate 4D representation, and propose a practical training paradigm that takes advantage of both generative adversarial networks (GANs) and diffusion models.
Our design stems from the observation that 4D GANs excel at bridging images and triplanes without supervision yet usually face challenges in handling diverse data distributions.
A robust 2D diffusion prior emerges as the solution, assisting the GAN in transferring its expertise across various domains.
The synergy between these experts permits the construction of a multi-domain image-triplane dataset, which drives the development of a general 4D avatar creator.
Extensive experiments suggest that our model, termed \method, is capable of producing high-quality 4D avatars with strong robustness to various source image domains.
The code, the data, and the models will be made publicly available to facilitate future studies.

\nonumfootnote{* This work is done partially when Hongyu is an intern at Ant Group.}%
\nonumfootnote{\S~Joint corresponding authors.}%

\end{abstract}

\begin{figure*}[t!]
    \begin{center}
        \includegraphics[width=1\linewidth]{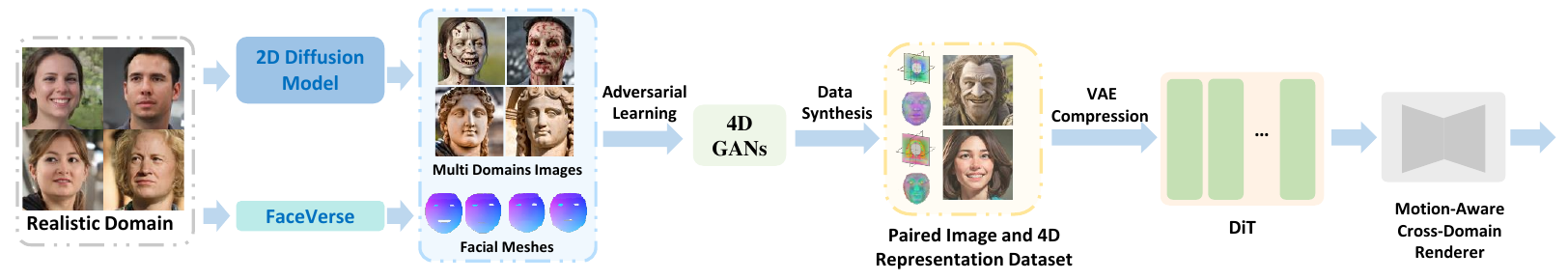}
    \end{center}
    \vspace{-1em}
    \caption{\textbf{The overall training pipeline of our method.} We first generate 2D images from different domains using a 2D diffusion model. These images are then used to train 4D GANs for each domain. Subsequently, the trained 4D GANs generate image-4D representation pairs across domains, which are used to train DIT and the rendering model.}
    \label{fig:overall}
    \vspace{-1em}
\end{figure*}
\section{Introduction}\label{sec:intro}
Avatarization (dynamic) from one single portrait image has become a fundamental ability of AI content generation. It enables the transfer of motion and expression from a source video to a digital avatar while preserving both motion accuracy and subject identity. This technology has broad applications in virtual reality, social media, gaming, and online education, facilitating efficient character production and enhancing interactive experiences in computer vision and computer graphics.

Studies on avatarization can mainly be categorized as 2D and 4D aspects. The 2D-based methods~\cite{siarohin2019animating,siarohin2019first,zakharov2019few,siarohin2021motion,zhao2022thin,guo2024liveportrait,zhang2023metaportrait} typically employ a self-supervised learning scheme, with monocular video stream data accompanied by facial landmarks or implicit motion representations~\cite{siarohin2019first}. 
More recently, the emergence of powerful generative models, such as diffusion models, which can handle various types of images, has further advanced the field. Some 2D methods~\cite{wei2024aniportrait,xie2024x,ma2024followyouremoji} incorporate prior knowledge from diffusion models (e.g., Stable Diffusion~\cite{rombach2021highresolution}), enabling them to effectively handle multi-style avatarization (e.g., cartoon, realistic)."
Despite achieving impressive results, these 2D methods fail to accurately represent 3D structures. Geometric distortion and content inconsistency often arise when the head pose undergoes significant rotation. Moreover, the iterative computation of diffusion models incurs substantial costs for generating each frame of animated videos, significantly increasing the overall computational burden.

On the other hand, 4D-based methods~\cite{ma2023otavatar, deng2024portrait4d, zhao2024invertavatar,li2023one,chu2024gagavatar} leverage neural rendering pipelines~\cite{mildenhall2021nerf,kerbl3Dgaussians} and 3DMM~\cite{FLAME:SiggraphAsia2017} for efficient avatarization where 3D geometric consistency is maintained across multiple viewpoints. During model inference, these models animate the image feature first then a camera pose to perform neural rendering of target view generations. Despite the demonstrated success, these methods suffer from a lack of 4D data from diverse domains. The human portrait animation is restricted to a limited domain and is difficult to generalize as that of 2D-based methods. 

\textit{``Having examined both 2D and 4D-based avatarization methods, we intuitively assume that if sufficient and well-suited 4D datasets covering diverse domains were available, it would be possible to develop a 4D avatarization approach for open-domain inputs using diffusion models.''} Recently, Rodin~\cite{wang2023rodin, zhang2024rodinhd}, a diffusion-based single-image-guided static avatar generation method, has demonstrated impressive performance in the synthetic digital domain. To train this model, a dataset of image-3D representation pairs was constructed using multi-view digital human data. Inspired by this, we believe that an appropriate 4D dataset for our method should consist of image-4D representation pairs spanning multiple domains.

In this work, we propose AvatarArtist, a diffusion-based 4D avatarization model.  It is challenging to obtain multi-view, multi-expression 4D captures to create image-4D representation pairs with a fitting process similar to Rodin. Therefore, we resort to synthetic data generation. Fortunately, we found that 4D GANs, such as Next3D~\cite{sun2023next3d, zhao2024invertavatar}, can greatly assist in this process. Specifically, Next3D proposed a parametric triplane 4D representation, which divides the traditional triplane~\cite{chan2022efficient} into dynamic and static components. The dynamic part is aligned with the 3DMM mesh~\cite{wang2022faceverse, FLAME:SiggraphAsia2017} in UV space, allowing expression changes through mesh rasterization and rendering. With Next3D, we can generate an unlimited amount of image-parametric triplane data simply by sampling, but only for single realistic domain due to the mode collapse issue of GAN. Hence we propose to finetune Next3D to efficiently obtain multiple GANs of diverse domains. While training Next3D only requires 2D images and their corresponding 3DMM meshes, achieving effective multi-domain fine-tuning demands diverse and comprehensive data coverage across various domains. To overcome this limitation, we leverage 2D diffusion models~\cite{rombach2021highresolution} to enrich the diversity of the training data.
Specifically, we utilize the SDEdit~\cite{meng2021sdedit} pipeline and landmark-guided ControlNet to transfer portrait images (e.g., FFHQ) from the realistic domain to other domains. This process ensures coherent pose and expression between the output and input 2D portraits, allowing us to reuse the 3DMM mesh of the 2D portrait from the realistic domain in non-realistic domains. Consequently, we can train 4D GANs for different domains and generate image-parametric triplane datasets across multiple domains. The entire data generation process combines the advantages of both diffusion models and GANs: diffusion models provide multi-domain data for the GAN, while the GAN transforms 2D images into 4D representations in an unsupervised manner.

Using this dataset, we could adopt the latent Diffusion Transformer (DiT)~\cite{Peebles2022DiT} to model its distribution. The process begins with training a VAE to compress triplanes into latent representations, followed by employing a DiT to generate latent guided by a single portrait image. Although the diffusion model is able to generate triplanes effectively, there are still two issues for rendering high-quality frames from these planes. First, not like Rodin which uses a simple MLP decoder for the digital domain, rendering triplanes from multiple diverse domains into high-quality images is much more challenging. Second, Parametric triplanes primarily focus on motion modeling but are less effective in preserving identity. Next3D employs a CNN to enhance identity preservation,  but we found the performance of CNN degrades significantly in the open domain. To address these, we introduce a motion-aware cross-domain renderer based on ViT~\cite{xie2021segformer} that incorporates features from the source image, improving cross-domain rendering ability and preserving the identity information.   Additionally, we use an implicit motion representation, similar to Portrait4D~\cite{deng2024portrait4d}, to avoid artifacts caused by mesh inaccuracies. Compared to baseline methods, our approach delivers superior quantitative and qualitative performance, offering high visual fidelity, accurate identity representation, and precise motion rendering.

\begin{figure*}[t!]
    \begin{center}
        \includegraphics[width=1\linewidth]{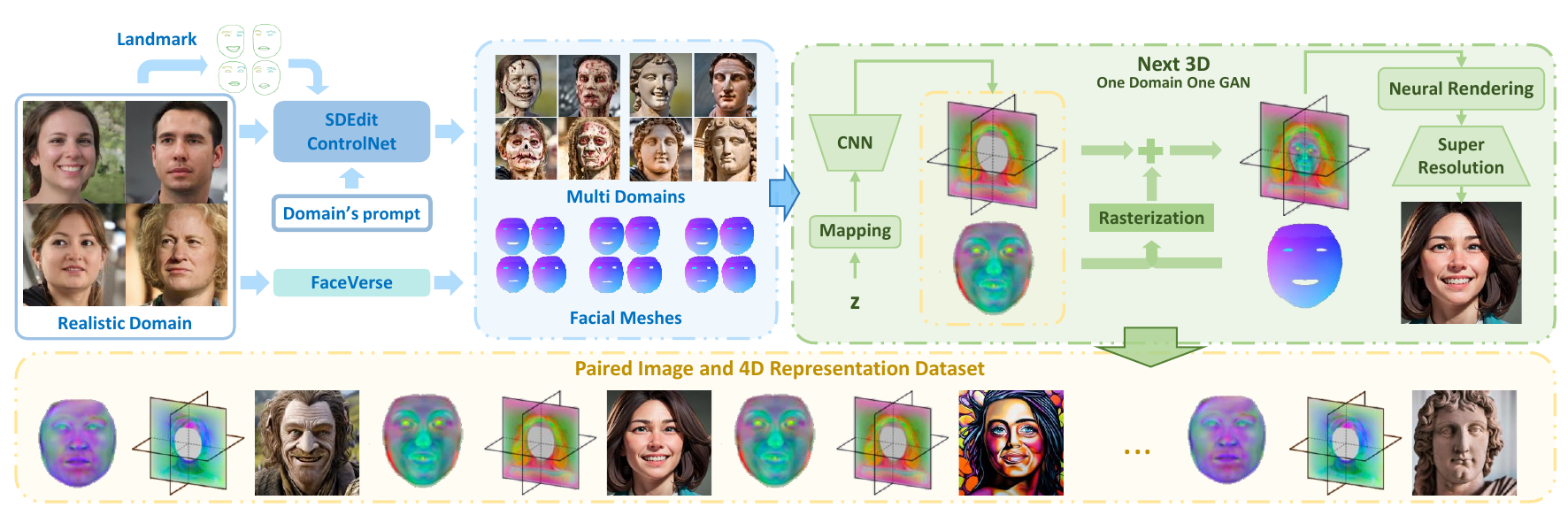}
    \end{center}
    \vspace{-1em}
    \caption{\textbf{The pipeline of dataset generation.}We use text prompts to transform images from the realistic domain to the target domain while ensuring pose and expression consistency with SDEdit~\cite{meng2021sdedit} and landmark-guided ControlNet~\cite{zhang2023adding}. This enables direct reuse of the original mesh, avoiding errors in non-realistic domain extraction. After domain transfer, we train 4D GANs to generate image-parametric triplane pairs, which serve as data for the next stage. The parametric triplane comprises dynamic and static components, with the dynamic region aligned to the mesh. }
    \label{fig:data_gen}
\vspace{-0.18in}

\end{figure*}

\section{Related Work}\label{sec:related}
We address one-shot, open-domain image-driven talking face generation, which synthesizes a talking head video from a single reference portrait and a sequence of driving expression images. This section provides a concise overview of previous talking head generation methods, broadly categorized into 2D talking face generation and 3D-aware talking portrait synthesis, along with a brief discussion on stylized 3D avatar generation.

\subsection{2D Talking Face Generation}
Great progress has been made in image-driven 2D talking head generation~\cite{siarohin2019animating,siarohin2019first,zakharov2019few,burkov2020neural,drobyshev2022megaportraits,siarohin2021motion,gong2023toontalker,yin2022styleheat,wang2023progressive, xu2024vasa,zhang2023metaportrait,ma2024followyourpose,liu2023human}. Numerous approaches harness the capabilities of Generative Adversarial Networks to synthesize high-fidelity talking head videos, most of which fall into the warping-then-rendering scheme. The identity features are first encoded from the reference image and then warped according to the driving signals, finally being rendered into a sequence of talking portraits. More specifically, various types of motion representation, such as landmarks~\cite{zakharov2019few,siarohin2019first}, depth~\cite{hong2022depth}, and latent code~\cite{burkov2020neural}, are exploited to deduce the warping field, ensuring that the synthesized portraits exhibit expressions and motions that faithfully correspond to the driving signals. With the advent of diffusion model-based image generation, several methods employ large pre-trained diffusion models to assist in the task of one-shot talking face generation. By leveraging the powerful prior of pre-trained diffusion models, recent methods~\cite{xie2024x, ma2024followyouremoji, wei2024aniportrait} have demonstrated that they possess strong generalization capabilities when handling various styles of reference portraits. However, due to a lack of understanding of three-dimensional structures, these 2D-based methods often exhibit obvious geometric distortions when handling larger head movements. Additionally, they lack the ability to control the viewpoint of the generated images with precision.

\subsection{3D-aware Talking Portrait Synthesis}

To achieve high geometric fidelity in synthesizing portraits with varying head poses, researchers have introduced intermediate 3D representations that capture facial geometry and pose, ensuring structural accuracy across viewpoints. A major breakthrough in novel view synthesis is Neural Radiance Fields (NeRF)\cite{mildenhall2021nerf, chan2022efficient, hong2022headnerf, yu2023nofa, li2023one, li2024generalizable, ma2023otavatar, zhuang2022mofanerf, trevithick2023real, chu2024gpavatar, ye2024real3d}, which enables precise 3D reconstructions with explicit camera control. NeRF has been widely adopted in 3D-aware one-shot talking head generation, enhancing 3D coherence and pose control for more natural outputs. More recently, GAGAvatar\cite{chu2024gagavatar} leveraged 3D Gaussian Splatting (3DGS) to accelerate generation while maintaining high quality.

However, most methods~\cite{wang2021one} rely on in-the-wild video data, making 3D learning from monocular videos highly ill-posed due to depth ambiguity, lighting variations, and facial occlusions. Some approaches incorporate 3D supervision from monocular 3D face reconstruction~\cite{danvevcek2022emoca,deng2019accurate,feng2021learning}, multi-view lab-captured videos\cite{hong2022headnerf,zhuang2022mofanerf}, or synthetic multi-view data\cite{deng2024portrait4d,deng2024portrait4dv2}. While these improve results, they are constrained by limited high-quality 3D data and training challenges. As a result, there remains no open-domain, one-shot 4D portrait generation method capable of generalizing across diverse conditions.

\subsection{Stylized Avatar Generation}
To generate avatars across different domains, some methods~\cite{kim2022datid3d, Abdal_2023_CVPR, kim2023podia, song2024texttoon, perez2024styleavatar, 10.1145/3641519.3657512, Lei_2024_CVPR, wan2024cad,bai2024real} use CLIP as a constraint or leverage diffusion models to generate reference images, which are then utilized to create stylized avatars based on text prompts. Additionally, StyleAvatar3D~\cite{zhang2023styleavatar3d} and Rodin~\cite{wang2023rodin,zhang2024rodinhd} collect domain-specific datasets to train generative models for stylized avatar synthesis. While these methods significantly advance stylized avatar generation, they do not focus on single-image-guided, animatable avatar creation. Meanwhile, other approaches~\cite{alanov2022hyperdomainnet, zhu2022one, patashnik2021styleclip, gal2022stylegan} employ CLIP as a constraint and text as guidance to fine-tune GAN models, enabling the generation of stylized portrait images that align with textual descriptions. Although these methods demonstrate strong manipulation capabilities for stylized portraits, they cannot directly generate avatars.

\section{Method}\label{sec:method}

We aim to develop a system that generates a 4D avatar from an open-domain image $I_s$, driven by the motion of a target individual $I_t$. The key to training a deep generative model for open-domain avatarization is a large-scale, high-quality dataset. In Sec.\ref{sec.3.1}, we introduce how GANs and image generation techniques help construct diverse and consistent training data. With this dataset, we use a latent Diffusion Transformer (DiT) to model the 4D distribution (Sec.\ref{sec.3.2}). To ensure accurate motion transfer while preserving the source identity, we further employ a motion-aware cross-domain renderer (Sec.\ref{sec.3.3}). The overall training pipeline is shown in Figure\ref{fig:overall}. Next, we detail each component.

\subsection{Data Curation from 4D GANs}\label{sec.3.1}

Benefiting from adversarial training, the recent GAN methods have demonstrated great potential in generating high-quality 4D avatars in an unsupervised manner using non-multiview images and 3DMM meshes only. Therefore, we would like to fully leverage this capability of GANs to curate 4D data. Nonetheless, the instability of GAN training easily caused mode-collapse, failing to cover the distribution of different modes. In this section, we will discuss how to properly use GAN to generate open-domain image-4D representation pairs data.

\noindent\textbf{Base GAN Model.} We select Next3D~\cite{sun2023next3d, zhao2024invertavatar} as our base GAN model for generating the 4D dataset, given its training efficiency and the proposed robust 4D representation (parametric triplane). Specifically, as shown in Figure~\ref{fig:data_gen}, given a randomly sampled latent code $z$, the mapping network translates $z$ into an intermediate latent vector, which will modulate conv layers of StyleGAN to generate a parametric triplane $\in  \mathbb{R}^{256 \times256 \times 4 \times 32}  $. The parametric triplane consists of two parts: a static component representing non-facial regions and a dynamic component aligned with the 3DMM mesh in UV space. During inference, given a specific mesh, the dynamic part is deformed through rasterization and combined with the static part to form a triplane with expressions. Then neural rendering and a super-resolution module are applied to generate the final image. We follow Next3D and use FaceVerse~\cite{wang2022faceverse} to extract the corresponding parametric mesh. 
\begin{figure}[t!]
    \begin{center}
        \includegraphics[width=1\linewidth]{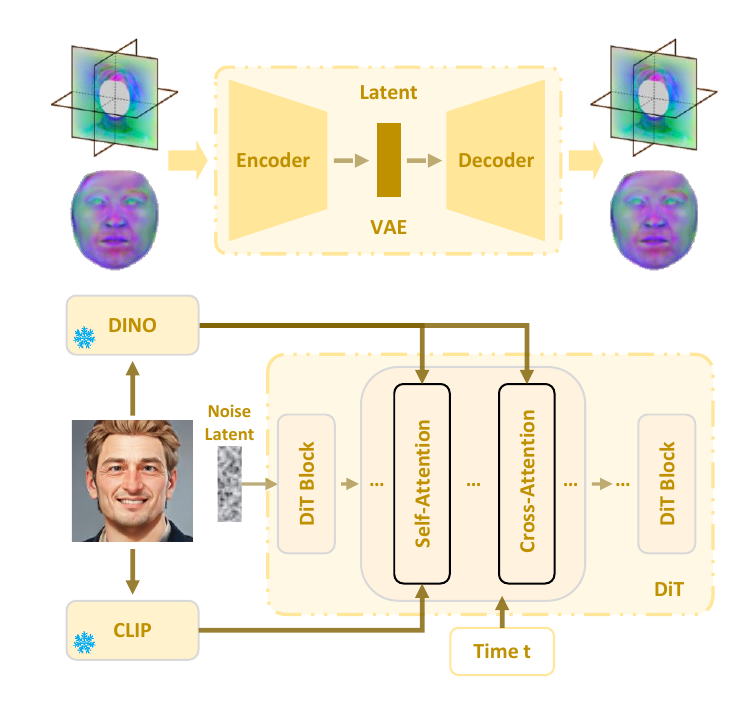}
    \end{center}
    \vspace{-1em}
    \caption{\textbf{The pipeline of DiT.} We first train a VAE to compress the parametric triplane into a latent space, and then train a DiT to denoise the noisy latent. We incorporate features from DINO \cite{caron2021emerging} and CLIP~\cite{radford2021learning} into the DiT to guide the generation process. }
    \label{fig:pipe_dit}
    \vspace{-0.6em}
\end{figure}
\noindent\textbf{Multi-Domain Tuning.} To train Next3D models across various domains, the first priority is to obtain diversified images from different domains and extract the corresponding 3DMM meshes. However, it is very challenging to accurately obtain 3DMM meshes for non-realistic portraits. To address this issue, we use the pre-trained 2D diffusion model to generate the target domain from realistic images given specific prompts, so that the 3D meshes from realistic domains can be re-used. Specifically, given a portrait from the realistic domain,  we add noise over this image with specific strength~\cite{meng2021sdedit} and then denoise it by StableDiffusion to generate high-quality and diversified images of the target domain under the guidance of prompts. However, corrupting the images with noise also raises challenges in maintaining the original expression. Hence, we also incorporate the facial landmark through ControlNet~\cite{zhang2023adding}, which provides the conditional signals of expression. As a result, the newly sampled image from the target domain could closely align with the pose and expression of the source realistic image, enabling the direct transfer of 3DMM mesh labels from the realistic domain to the target domain.

Through the pipeline mentioned above, we totally collected image data from 28 different domains (anime, lego, etc.) transferred from FFHQ~\cite{karras2019style}. To ensure the efficiency of the data pipeline and avoid model collapse, for each domain, we generate 6,000 images only and use this data to finetune independent GAN from the Next3D model trained with FFHQ~\cite{karras2019style}. We follow the DATID-3D~\cite{kim2022datid3d} to use the ADA loss and density regularization loss to guarantee the diverse content generation ability of GAN during tuning.

\noindent\textbf{Data Synthesis.} We utilize the trained multiple 4D GANs to build two datasets. 1) The image-parametric triplane paired dataset. We randomly sampled poses and meshes of portrait images from the FFHQ dataset, which are then fed into the 4D GAN with a random \( z \)  to generate images in different identities along with corresponding triplanes. For each domain, we generated 20K samples, resulting in a total of 20K $\times$ 28 = 560K image-triplane pairs. 2) The multi-view, multi-expression image-parametric triplane dataset. These data assist in learning our motion-aware cross-style renderer. We generate both static and dynamic components following the portrait4D~\cite{deng2024portrait4d}. The dynamic data, which are responsible for head reenactment, consists of synthetic identities with multiple expressions per subject and varying camera poses for each expression. The static data, on the other hand, are employed to enhance the generalizability of 3D reconstruction and contain only a single expression per identity, also with varying camera poses. Expressions (meshes) in the dynamic dataset are sampled from the VFHQ dataset~\cite{xie2022vfhq}, while those in the static dataset are sampled from the FFHQ dataset. All camera poses are sampled from FFHQ.

\subsection{4D Generation}\label{sec.3.2}
Recently, the latent diffusion model has shown great potential in modeling complex data distributions like images~\cite{podell2023sdxl}, videos~\cite{polyak2024movie}, and triplanes~\cite{wang2023rodin, Chan2021, Shue_2023_CVPR}. Following this trend, this section will depict how we leverage latent diffusion for 4D generation. As shown in Figure~\ref{fig:pipe_dit}, we will first introduce a triplane VAE to compress the triplane representations into a latent space, followed by training an image-conditioned DiT~\cite{Peebles2022DiT}. All the data used for training was sourced from our curated datasets.

\noindent\textbf{Triplane VAE.}
For training efficiency, the DiT ~\cite{Peebles2022DiT} is trained in a compact latent space by default. To achieve this, we propose a triplane variational autoencoder (VAE)~\cite{kingma2013auto} to obtain the latent code of the triplane representations. Specifically, our VAE compresses the triplane  $\in  \mathbb{R}^{256 \times256 \times 4 \times 32}  $   to latent $z_t$ $\in  \mathbb{R}^{64 \times 64 \times 4 \times 8}  $.  To optimize the VAE model, we compute the $\mathcal{L}_{1}$ loss between the reconstructed planes and input triplanes. Meanwhile, we also get depth and rendered images to calculate $\mathcal{L}_{1}$ and LPIPS losses, respectively. We did not apply adversarial loss since we found it introduced training instability. For more details, please refer to our supplementary material.

\noindent\textbf{Image Guided Diffusion Transformer.} We follow the Direct3D~\cite{wu2024direct3d} and PixArt-$\alpha$~\cite{chen2023pixartalpha}to build our image-guided DiT. For the noised latent $z_t$ we flatten it to a sequence and send it to the DiT as input. We separately extract the semantic and detailed information from conditional images and inject them into each DiT block. For semantic information, we use CLIP~\cite{radford2021learning} to extract the image's semantic embeddings, which are then integrated with the model via cross-attention. To capture fine-grained details, we employ the DINO~\cite{oquab2023dinov2} to extract image tokens. In each DiT block, we concatenate the image tokens with the flattened $z_t$ and feed them into a self-attention layer to model the intrinsic relationship between the image tokens and $z_t$. During training, we leverage the objective of IDDPM~\cite{nichol2021improved} and predict the noise and variance at each time step t. We also randomly drop the conditional image with a probability of 10\% to enable classifier-free guidance~\cite{ho2022classifier} during inference.

\subsection{Motion-Aware Cross-Domain Renderer}\label{sec.3.3}

In the rendering process of Next3D, a CNN refines the rasterized parametric triplane to protect the identity information and eliminate identity leakage caused by rasterization. However, we found that this rendering approach fails to achieve acceptable quality in our setting (see Figure~\ref{fig:ab_model}). Since our parametric triplane is generated from different domains, a simple rendering network cannot effectively resolve identity leakage across various domains. Additionally, inaccurate mesh extraction sometimes leads to mismatched expressions in the generated results.

\begin{figure}[t!]
    \begin{center}
        \includegraphics[width=1\linewidth]{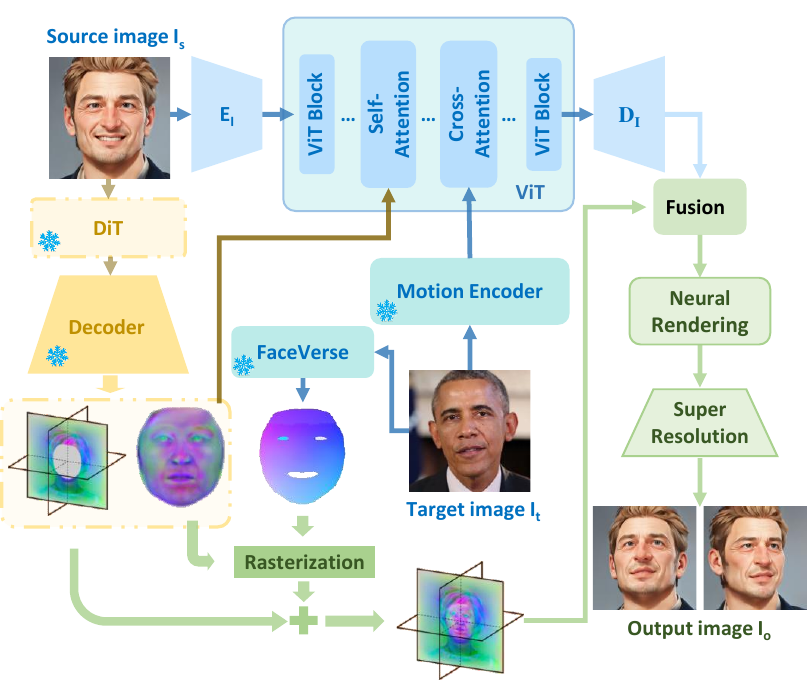}
    \end{center}
    \vspace{-0.7em}
    \caption{\textbf{The pipeline of motion-aware cross-domain renderer.} We use an encoder to extract the feature from the source image. This feature is sent to a  ViT to predict results under the guidance of generated parametric triplane and motion embedding. Finally, a decoder processes the output of the ViT and fuses it with the results of rasterization to produce the final output. }
    \label{fig:pipe_refine}
    \vspace{-0.6em}
\end{figure}

 To address these issues, we propose a motion-aware cross-domain renderer. As shown in Figure~\ref{fig:pipe_refine}, we first employ an encoder $E_I$ to extract features from the source images, which are subsequently fed into a Vision Transformer (ViT) model~\cite{xie2021segformer}. In the ViT model, we inject the parametric triplane generated by DiT into the self-attention mechanism, which aims to neutralize facial expressions and canonicalize poses inspired by ~\cite{deng2024portrait4d}, thereby eliminating expression-specific information from the source image. Then, we change the expression with motion embedding~\cite{wang2023progressive}  by injecting it with cross attention.  This embedding is an implicit representation without spatial information, thus preventing identity leakage. The output of the ViT is decoded to match the resolution of the rasterized parametric triplane, after which it is fused with the triplane. Finally, we apply volumetric rendering followed by super-resolution techniques to generate the final output $I_o$.  The $I_o$, rendered from a novel camera viewpoint, preserves the identity features from the source image  $I_s$ meanwhile following the facial expression of the target image  $I_t$. For training this model, we adopt the loss terms following the  ~\cite{10.1145/3641519.3657478, deng2024portrait4d, trevithick2023real}.  For more details, please refer to our supplementary material.

 \begin{figure*}[t]
\begin{center}
\includegraphics[width=1\linewidth]{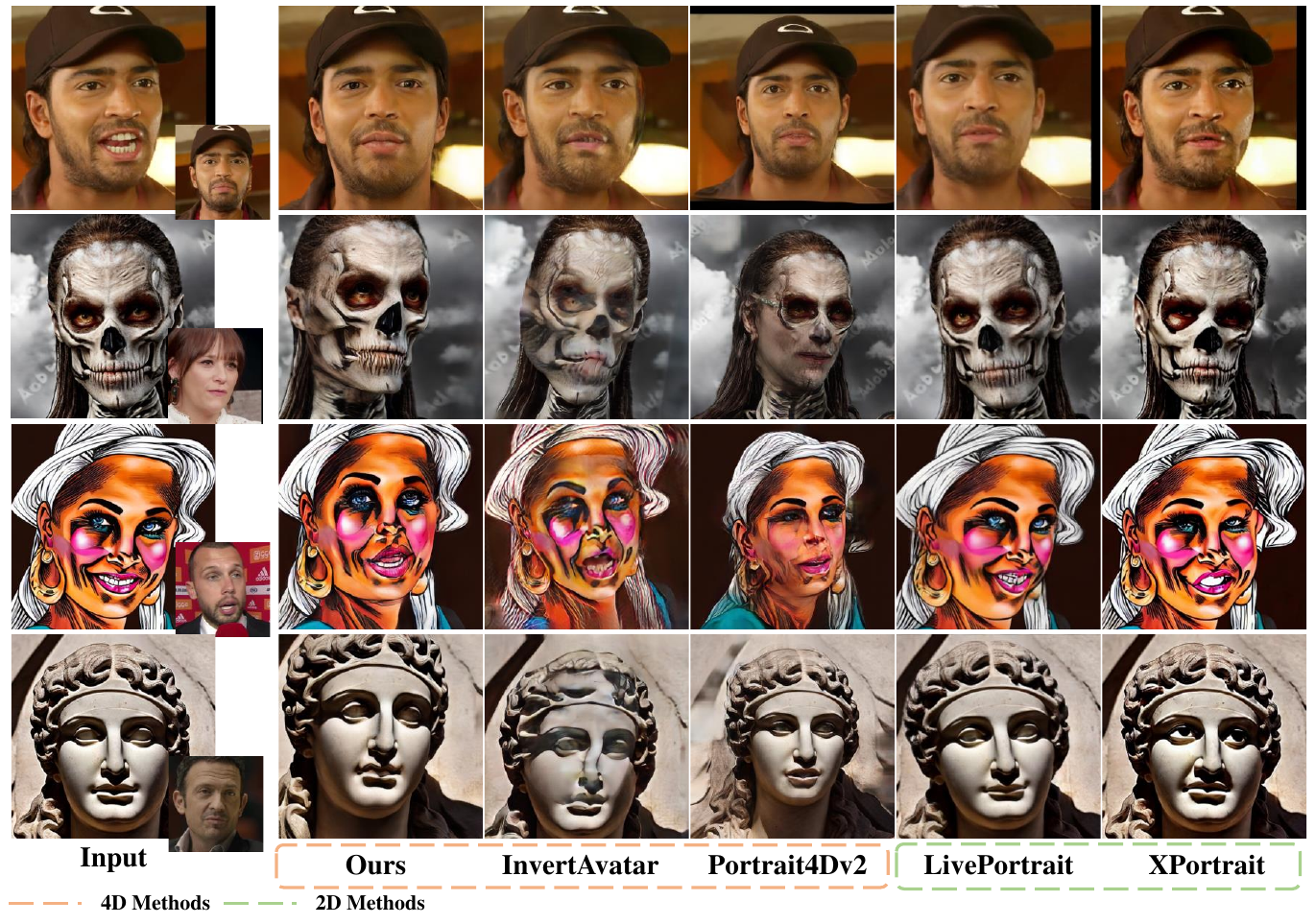}
\caption{Qualitative comparison with SOTA methods.  The leftmost column in the figure shows the input images, with the bottom-right corner representing the target image. The first row displays the results of self-reenactment, while the following three rows show the results of cross-reenactment. It can be observed that our method achieves superior performance in terms of expression and pose consistency, as well as identity preservation.   }
\label{fig:compare}
\vspace{-0.1in}

\end{center}
\end{figure*}

\begin{table*}[t]
\centering
\caption{Quantitative evaluation of state-of-the-art methods and our approach on the VFHQ dataset~\cite{xie2022vfhq}. 
For self-reenactment, both the source and target images are from the VFHQ dataset.
For cross-ID reenactment, the source images are generated from different domains, while the target motions are from VFHQ.
${\downarrow}$ indicates lower is better while ${\uparrow}$ indicates higher is better.
\textbf{\textcolor{lightred}{Red}} highlights the best result, and \textbf{\textcolor{lightblue}{Blue}} highlights the second-best result.}
\resizebox{0.8\linewidth}{!}{
\begin{tabular}{l|lllll|llll}
\toprule
\multicolumn{1}{c}{\multirow{2}{*}{Method}} &  \multicolumn{5}{c}{Self reenactment}  & \multicolumn{4}{c}{Cross reenactment} \\ 
\cline{2-10} 
 \multicolumn{1}{c}{} & LPIPS $\downarrow$ &  FID $\downarrow$ & ID $\uparrow$ & AED $\downarrow$  & APD $\downarrow$ & FID $\downarrow$ & CLIP $\uparrow$ & AKD $\downarrow$ & APD $\downarrow$\\
 \midrule
 LivePortrait~\cite{guo2024liveportrait}       &  0.27                     & \textbf{\textcolor{lightred}{46.49}}              &\textbf{\textcolor{lightblue}{0.65}}                  &\textbf{\textcolor{lightblue}{0.025}}               &\textbf{\textcolor{lightred}{4.28}}                 & 100.3                   &\textbf{\textcolor{lightred}{0.91}}                    & 4.92                     &  139.35                  \\
 XPortrait~\cite{xie2024x}                     &  0.31                     & 60.29               &0.63                 & 0.036              &18.07                  &\textbf{\textcolor{lightred}{78.6}}           & \textbf{\textcolor{lightblue}{0.89}}                          & 10.67                      &  237.4\\ 
 \midrule
 InvertAvatar~\cite{zhao2024invertavatar}      &  0.42                   & 84.71               &0.32                  & 0.049                &  15.58                 & 194.7                     & 0.64                        & 20.78                       & 134.9               \\
 Portrait4Dv2~\cite{deng2024portrait4dv2}      &  0.29                    & 66.60               & 0.58                  &0.034                & \textbf{\textcolor{lightblue}{ 5.08}}          &140.5                     &  0.75                       &7.13                        & 63.3        \\
\midrule
Ours                                           & \textbf{\textcolor{lightred}{0.26}}            &\textbf{\textcolor{lightblue}{52.62}}     &\textbf{\textcolor{lightred}{0.69}}          &\textbf{\textcolor{lightblue}{0.032}}       & 11.72                 &\textbf{\textcolor{lightblue}{89.3}}                     & 0.84                        &\textbf{\textcolor{lightred}{2.58}}                & \textbf{\textcolor{lightblue}{52.3}}       \\ 
\bottomrule
\end{tabular}}
\label{tab:experiment}
\vspace{-0.08in}

\end{table*}

\begin{table}[t]
\centering
\caption{Ablaiton study on the FFHQ~\cite{karras2019style} dataset. The source images are generated from different domains, while the target images are from FFHQ. The Next3D rendering means replacing our render model with simple CNN.}
\resizebox{0.85\linewidth}{!}{
\begin{tabular}{l|llll}
\toprule
\multicolumn{1}{c}{\multirow{2}{*}{Method}}  & \multicolumn{4}{c}{Cross reenactment} \\ 
\cline{2-5} 
 \multicolumn{1}{c}{} & FID $\downarrow$ & CLIP $\uparrow$ & AKD $\downarrow$ & APD $\downarrow$\\
 \midrule
Next3D render    &130.72            &   0.73            &   5.89&                                                42.93             \\
\midrule
Ours &             \textbf{68.69} &   \textbf{0.86}&   \textbf{2.56}&       \textbf{40.89}            \\ 
\bottomrule
\end{tabular}
}
\label{tab:ablation}
\vspace{-0.05in}

\end{table}

\section{Experiments}\label{sec:exp}
In this section, we first illustrate our implementation details. Then, we compare our method with existing methods qualitatively and quantitatively. We compare our approach with both 2D and 4D reenactment methods. Specifically, we include comparisons with 2D techniques such as LivePortrait \cite{guo2024liveportrait} and XPortrait \cite{xie2024x}, as well as 4D methods like InvertAvatar \cite{zhao2024invertavatar} and Portrait4Dv2~\cite{deng2024portrait4dv2}. Finally, an ablation study validates the effectiveness of our contributions.  More results are provided in the supplementary files. 

\subsection{Implementation Details}

During the training of the Next3D, we extract facial poses and corresponding 3DMM meshes from the FFHQ dataset using FaceVerse~\cite{wang2022faceverse}. All domains are fine-tuned based on a GAN pre-trained on the FFHQ dataset, with each domain iterating over a total of 300K images. For VAE training, we adopt the same training framework as the VAE used in Stable Diffusion. We utilize the AdamW optimizer~\cite{loshchilov2019adamw} with a learning rate of 1e-4. The VAE model is trained on an NVIDIA A100 (80G) GPU for 100K steps with a batch size of 32. Our diffusion model follows the network configuration of DiT-XL/2~\cite{chen2023pixartalpha, wu2024direct3d, Peebles2022DiT}, consisting of 28 layers of DiT blocks. The diffusion model is trained with 1000 denoising steps using a linear variance scheduler. We employ the AdamW optimizer with a learning rate of 1e$^{-4}$ and train the model for 800K steps. During inference, we use 19 steps of the DPMSolver~\cite{lu2022dpm}, with a guidance scale set to 4.5. For the motion-aware cross-domain renderer, we train on a total of 12 million images across all domains. For more details, please refer to the supplementary materials.

\subsection{Qualitative Results} 
As shown in Figure~\ref{fig:compare}, we present a visual comparison of the results of self-reenactment and cross-reenactment tasks. The first column contains the input images, with the bottom-right corner showing the target image and the larger images representing the source images.  The first row displays the self-reenactment results. We observe that InvertAvatar exhibits noticeable artifacts, while XPortrait shows misalignment in pose compared to the target. Although Portrait4D and LivePortrait achieve relatively good results, there are inconsistencies in the expression, particularly in the eye and mouth regions, when compared to the target. In contrast, our method produces more consistent results, achieving better alignment with the target in both pose and expression. 

For the cross-reenactment, our source images are from non-realistic domains, while the target expressions are extracted from real-human domains. We observe that both InvertAvatar and Portrait4D struggle to handle portraits that significantly differ from real-human domains effectively. InvertAvatar tends to exhibit severe geometric distortions and fails to adequately animate the source image. Portrait4D, on the other hand, suffers from identity leakage and generates content that lacks precision. While 2D-based methods preserve the identity of the input image, they fail to ensure that the pose aligns with the target image. In contrast, our method demonstrates exceptional performance when handling non-realistic domains, achieving good accuracy in both expression and pose consistency, as well as identity preservation.

\subsection{Quantitative Results} The quantitative results are summarized in Table~\ref{tab:experiment}. We evaluate our method on 100 VFHQ video clips\cite{xie2022vfhq} through self-reenactment and cross-reenactment tests. For self-reenactment, the source image is either the first frame or a random intermediate frame from the video, while for cross-reenactment, we use 50 source images from different domains with target images from VFHQ. To assess image quality, we compute LPIPS\cite{zhang2018unreasonable} and FID\cite{heusel2017gans}. Identity consistency is measured using the ID metric\cite{deng2019arcface} for self-reenactment and CLIPScore\cite{radford2021learning} for cross-reenactment, as the ID metric is unreliable for non-human domains. Expression accuracy is evaluated with Average Expression Distance (AED)\cite{lin20223dganinversion} for self-reenactment and Average Keypoint Distance (AKD)\cite{lugaresi2019mediapipe} for cross-reenactment, as 3DMM struggles with non-realistic humans. Additionally, Average Pose Distance (APD) is used to assess pose consistency, with pose information extracted using\cite{10477888}.
As shown in Table~\ref{tab:experiment}, our method performs slightly worse than 2D approaches in self-reenactment, but remains comparable while surpassing 4D methods in overall effectiveness. In cross-reenactment, although 2D methods better preserve identity, our approach achieves higher accuracy in capturing pose and expression, demonstrating the advantages of 4D-based techniques.

\subsection{Ablation Study}
We analyze the impact of different data generation pipelines and the performance of each module in our model.
 \begin{figure}[t]
\begin{center}
\includegraphics[width=1\linewidth]{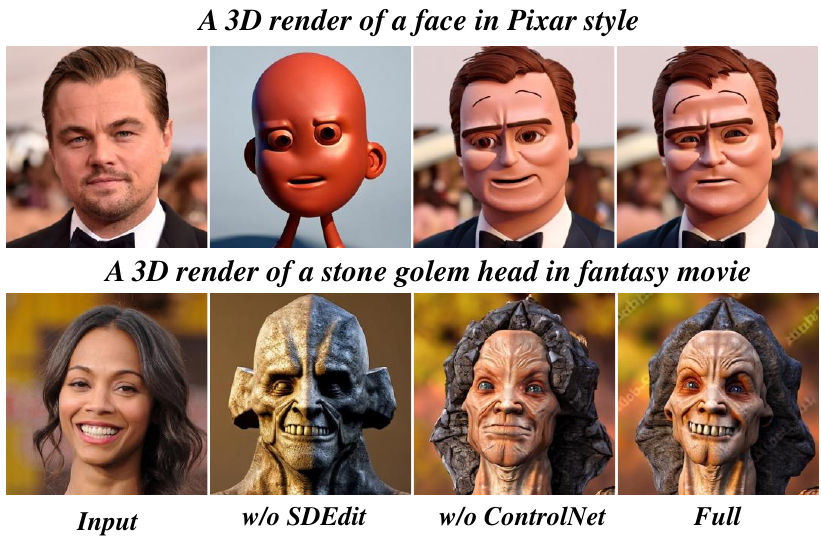}
\caption{ Visualization of ablation study on data generation methods. It is only when combining SDEdit and ControlNet that we can ensure the generated images retain both the same expression and pose as the original images. The corresponding prompts are shown above images.}
\vspace{-0.3in}

\label{fig:ab_sde_control}
\end{center}
\end{figure}

 \begin{figure}[t]
\begin{center}
\includegraphics[width=1\linewidth]{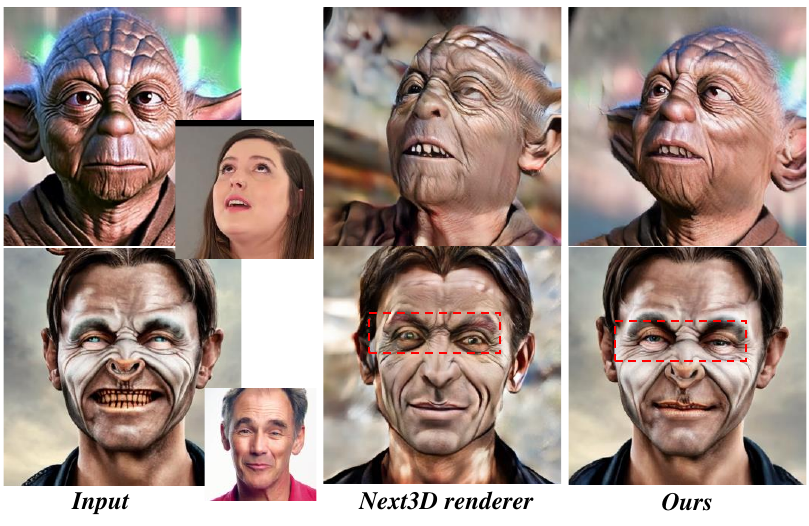}
\caption{Visualization of ablation study on motion-aware cross-domain renderer. The Next3D rendering approach involves using a CNN as a replacement for our render model. }
\vspace{-0.3in}
\label{fig:ab_model}
\end{center}
\end{figure}

{\flushleft \bf Effectiveness of Different Data Generation Methods.}As shown in Figure~\ref{fig:ab_sde_control}, the leftmost column presents the input images, all from the realistic domain. Without SDEdit, ControlNet provides some control over expressions, but the generated results still deviate significantly from the originals (w/o SDEdit). When using only SDEdit without ControlNet, the results preserve the pose, but the expressions still show noticeable discrepancies (w/o ControlNet). By combining ControlNet and SDEdit, we achieve images that maintain both the expression and pose of the original, while shifting entirely to a different domain (Full). This enables the reuse of 3DMM data from the realistic domain to train 4DGANs in various domains.

{\flushleft \bf Effectiveness of Models.} 
We design the motion-aware cross-domain renderer to better capture the appearance information from the original image, thereby enhancing fidelity. Additionally, since 3DMM is not perfectly accurate, we incorporate motion embedding to assist in the animation process. As shown in Figure~\ref{fig:ab_model}, we replaced our renderer with a CNN similar to the one used in Next3D. The results exhibited significant identity leakage (i.e., facial artifacts resembling the target subject's mesh), and the generated expressions did not accurately match the target (e.g., eye regions in the second row). In contrast, our method better preserves the source identity, and the implicit motion embedding effectively corrects motion inaccuracies from the mesh. The corresponding quantitative metrics in Table~\ref{tab:ablation}show that our approach outperforms all compared methods across all evaluated metrics.

\section{Conclusion}\label{sec:conclusion}
We propose AvatarArtist, a 4D avatar generation model for open-domain inputs. We use a parameterized triplane as a 4D representation and employ 4D GANs to build an open-domain image-triplane dataset. Specifically, a 2D diffusion model generates images from various domains, which train domain-specific 4DGANs to produce data for our model. Our model consists of DiT and a motion-aware cross-domain renderer. DiT converts the input image into a parameterized triplane, while the renderer refinement module synthesizes and optimizes results.
Experiments show that AvatarArtist effectively handles open-domain inputs, successfully transferring target motion while preserving appearance and geometric consistency.

\section{Acknowledgment}
This project was supported by the National Key R\&D Program of China under grant number 2022ZD0161501.

{
\small
\bibliographystyle{ieeenat_fullname}
\bibliography{ref.bib}
}

\appendix
\renewcommand\thesection{\Alph{section}}
\renewcommand\thefigure{S\arabic{figure}}
\renewcommand\thetable{S\arabic{table}}
\renewcommand\theequation{S\arabic{equation}}
\setcounter{figure}{0}
\setcounter{table}{0}
\setcounter{equation}{0}
\setcounter{page}{1}
\maketitlesupplementary

\section*{Appendix}
In the supplementary materials, we first discuss the limitations of our proposed method (Sec.~\ref{appendix:limitation}). Following that, we explore another 4D representation, providing a detailed analysis of the parametric triplane (Sec.~\ref{appendix:4DRepresentation}).We then explored different model architectures to validate the superiority of our DIT + render approach.  We  provide additional implementation details, including the domains used during training, the specific training procedures for each model, and other relevant training configurations (Sec.~\ref{appendix:implementation}). We provide additional comparisons and visual results to further demonstrate the effectiveness of our method (Sec.~\ref{appendix:results}). Last but not least, we present more results in the \textcolor{orange}{supplementary video}.

\section{Limitations}\label{appendix:limitation}
While our method can handle inputs from various domains and generate high-fidelity avatars, it does not adequately separate the head region from the background, nor does it decouple neck rotation from the camera pose, which limits the realism of the final results.  The 4D representation we employ uses a mesh as the primary driving signal. Although we incorporate motion embeddings as a supplementary motion signal, the process of obtaining the mesh is both time-consuming and imprecise, which adversely affects the overall efficiency and accuracy of the avatar generation.

\section{Exploration of the 4D Representation}\label{appendix:4DRepresentation}

 \begin{figure}[t]
\begin{center}
\includegraphics[width=0.96\linewidth]{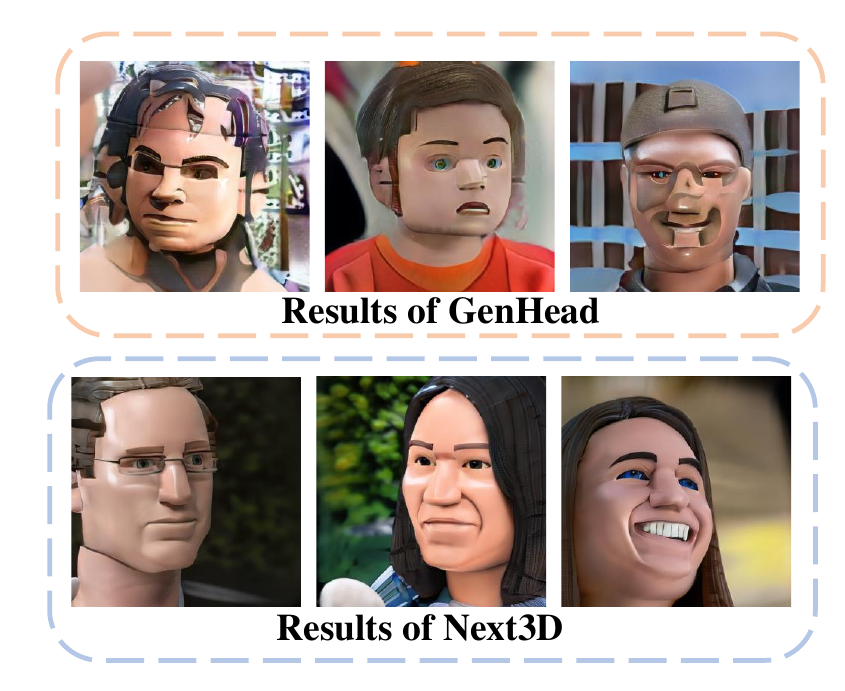}
\caption{Visualization of generation results of different 4D GANs, including Next3D~\cite{sun2023next3d} and GenHead~\cite{deng2024portrait4d}, on the unrealistic domain. We use the domain of Lego here. GenHead tends to produce artifacts, whereas Next3D achieves much better results, generating more plausible content. }
\vspace{-0.17in}
\label{fig:sup_4DGAN}
\end{center}
\end{figure}

In Portrait4D~\cite{deng2024portrait4d}, a 4D GAN (GenHead) based on a deformation field representation~\cite{Park_2021_ICCV} achieved impressive generative results. Specifically, the GenHead  $G$ consists of a part-wise triplane generator $G_{ca}$ for synthesizing the canonical triplane and a part-wise deformation field $D$ for morphing the canonical head. It generates the 3D deformation field based on FLAME~\cite{FLAME:SiggraphAsia2017} expression coefficients and synthesizes the canonical triplane using the shape parameter from FLAME. During inference, the canonical triplane can be driven by applying the deformation field to compute the offset for each point in the triplane with the corresponding Flame parameters. 

This canonical tri-plane and deformation field can also form a type of 4D representation. However, it is not suitable for our task. First, the deformation field changes according to different facial expressions, making it an unstable representation. In contrast, our representation only varies based on the subject's identity, ensuring consistency across different expressions for the same individual. Additionally, we found that GenHead does not perform well in open-domain generation. We suspect that this representation requires highly precise canonical space modeling, which is particularly challenging for non-realistic domains. In contrast, NeTX3D’s representation focuses more on motion modeling while delegating identity preservation to a separate CNN. Compared to GenHead, this representation is more implicit and better suited for generating characters across different domains. (See Figure~\ref{fig:sup_4DGAN}).

 \begin{figure}[t]
\begin{center}
\includegraphics[width=1\linewidth]{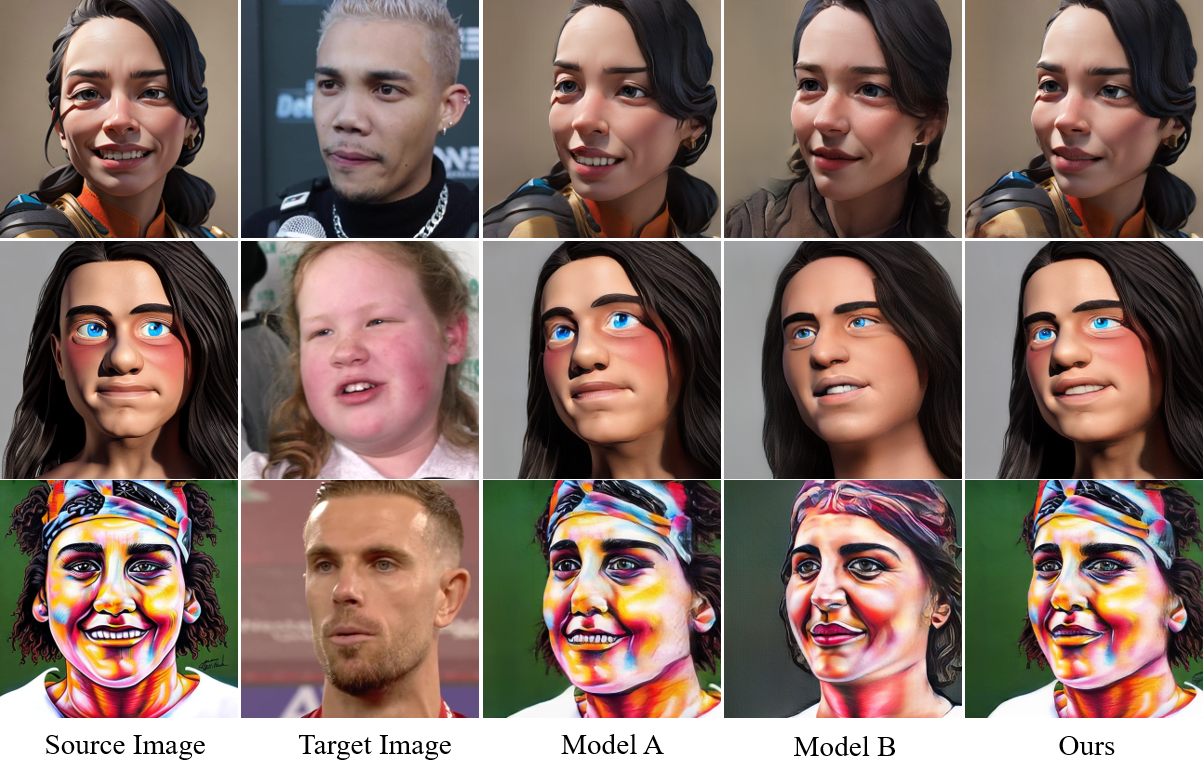}
\caption{Visualization of different model results. Model A and Model B are two different end-to-end method which not use the Dit. For more details,  please refer the Sec.~\ref{appendix:modeldesign}.} 
\vspace{-0.17in}
\label{fig: model design}
\end{center}
\end{figure}

\section{Effectiveness of model design}\label{appendix:modeldesign}
To demonstrate the clear effectiveness of using a DiT model for triplane generation, we conduct experiments comparing it with two feedforward approaches, as illustrated in Fig.~\ref{fig: model design}. Model A, similar to Portrait4D, uses only 4D RGB data, preserving identity well but struggling with motion transfer due to the absence of a unified 4D representation and limitations of cross-attention for cross-domain motion retargeting. Model B, which operates without cross-attention, uses an encoder-decoder to convert input images into parametric triplanes and a ViT decoder to refine animated features. While effective at transferring expressions, the encoder-decoder based feedforward model fails to reconstruct accurate triplanes, leading to identity loss and making it more challenging for the ViT decoder to bridge the identity gap. In contrast, similar to VASA-1~\cite{xu2024vasa}, our diffusion + renderer pipeline leverages the target parametric triplanes fitting ability of a powerful generative model. This enables our method to simultaneously maintain both motion and identity, achieving the highest quality results.

\section{More implementation details}\label{appendix:implementation}

\subsection{Training Domains}
As mentioned in our main paper, we used 28 domain images during training, including the original realistic domain.  We categorize our domains into two types. The first type uses the official Stable Diffusion 2.1 model~\cite{rombach2021highresolution} as the generative model. For this type, the text prompts used are shown in Table~\ref{appendix:sd}, and we generate images in 20 different domain styles, with 6,000 images per domain. The second type, as shown in Table~\ref{appendix:c}, utilizes third-party models in Civitai~\cite{Civitai} as the generative models, where each model corresponds to a specific style. For these models, the same text prompt is used across all models, and we set the prompt as "masterpieces, portrait, high-quality".
\begin{table*}[t]
\caption{List of full-text prompts corresponding to each domain. The images for these domains were generated using SD-V1.5 as the base model, in combination with corresponding prompts.}
\centering
\begin{tabular}{clclccl}
\hline
\multicolumn{3}{c}{Concise Name of Domain} &  & \multicolumn{3}{c}{Full text prompt} \\ \hline
                       & Pixar          &   &  &               &  a 3D render of a face in Pixar style             &     \\
\multicolumn{1}{l}{}   & Lego           &   &  &               &   a 3D render of a head of a lego man 3D model            &      \\
                       &Greek statue    &   &  &               &   a FHD photo of a white Greek statue              &      \\
                       & Elf            &   &  &               &   a FHD photo of a face of a beautiful elf with silver hair in live action movie            &      \\
                       &Zombie          &   &  &               &  a FHD photo of a face of a zombie             &      \\
                       &Tekken          &   &  &               &  a 3D render of a Tekken game character            &      \\
                       &Devil          &   &  &               &  a FHD photo of a face of a devil in fantasy movie           &      \\
                       &Steampunk      &   &  &                &  Steampunk style portrait, mechanical, brass and copper tones          &      \\
                       &Mario          &   &  &                & a 3D render of a face of Super Mario    &      \\ 
                       &Orc            &   &  &                &  a FHD photo of a face of an orc in fantasy movie          &      \\
                       &Masque            &   &  &                &  a FHD photo of a face of a person in masquerad          &      \\
                       &Skeleton           &   &  &                &  a FHD photo of a face of a skeleton in fantasy movie          &      \\
                       &Peking Opera           &   &  &                &  a FHD photo of face of character in Peking opera with heavy make-up         &      \\
                       &Yoda         &   &  &                &  a FHD photo of a face of Yoda in Star Wars         &      \\
                       &Hobbit         &   &  &                &  a FHD photo of a face of Hobbit in Lord of the Rings         &      \\
                       &Stained glass         &   &  &                &  Stained glass style, portrait, beautiful, translucent         &      \\
                       &Graffiti        &   &  &                &  Graffiti style portrait, street art, vibrant, urban, detailed, tag        &      \\
                       &Pixel-art        &   &  &                &  pixel art style portrait, low res, blocky, pixel art style        &      \\
                        &Retro          &   &  &                &  Retro game art style portrait, vibrant colors       &      \\
                        &Ink          &   &  &                &  a portrait in ink style, black and white image      &      \\
\hline
\label{appendix:sd}
\end{tabular}
\end{table*}

\begin{table*}[t]
\caption{List of models used for each domain. The images for these domains were generated using specific models as base models. All models were sourced from Civitai~\cite{Civitai}, an AI-Generated Content (AIGC) social platform. }
\centering
\begin{tabular}{clclccl}
\hline

\multicolumn{3}{c}{Concise Name of Domain} &  & \multicolumn{3}{c}{Model Name} \\ \hline
                       & 3D-Animation          &   &  &               &  3D Animation Diffusion-V1.0           &     \\
\multicolumn{1}{l}{}   & Toon           &   &  &                      &   ToonYou-Beta6            &      \\
                       &AAM    &   &  &                               &   AAM Anime Mix              &      \\
                       & Counterfeit            &   &  &               &   Counterfeit-V3.0            &      \\
                       &Pencil         &   &  &                       &  Pencil Sketch             &      \\
                       &Lyriel          &   &  &                      &  Lyriel-V1.6            &      \\
                       &XXM          &   &  &                          &  XXMix9realistic           &      \\ \hline
\end{tabular}
\label{appendix:c}
\end{table*}

\subsection{4D GAN}
The 4D GANs (Next3D) for different domains were fine-tuned from the original FFHQ GAN. Similar to DATID-3D~\cite{kim2022datid3d}, the training was stopped once the GAN had seen 200,000 images. We set the batch size to 32 and used 8 A100 GPUs to fine-tune the model for 2 hours.  A learning rate of 0.002 was used for both the generator and discriminator. For the discriminator’s input, we applied image blurring, progressively reducing the blur degree as described in~\cite{chan2022efficient,Karras2021}, and we did not employ style mixing during training. We used the ADA loss combined with R1 regularization, with the regularization coefficient set to $\lambda = 5$. Additionally, the strength of the density regularization was set to $\lambda_{\text{den}} = 0.25$.

 \begin{figure*}[t]
\begin{center}
\includegraphics[width=0.96\linewidth]{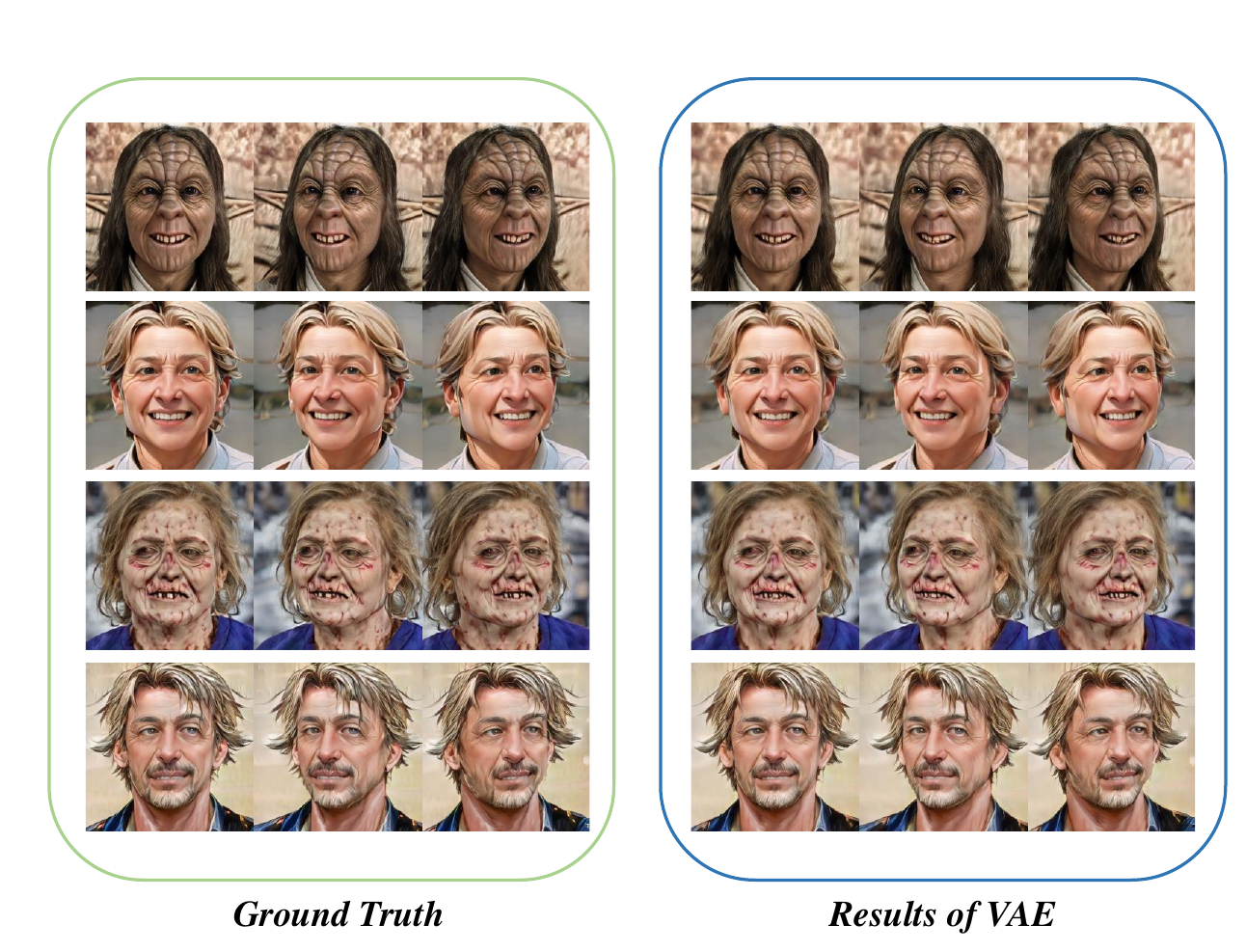}
\caption{Visualization of reconstruction results of our VAE. The domain is Yoda, 3D-Animation, Zommbie, and   Counterfeit, respectively. The ground truth images are generated with the Next3D. }
\vspace{-0.17in}
\label{fig:sup_vae}
\end{center}
\end{figure*}
\subsection{VAE}
 We follow the LVDM~\cite{he2022latent} and use a lightweight 3D autoencoder as our VAE. This VAE consists of an encoder $E$ and a decoder $D$. Both the encoder and decoder comprise multiple layers of 3D convolutions.   During training, we render the parametric triplane to obtain both depth maps and rendered images, and compute the $L_1$ and LPIPS losses separately. We also add a KL divergence loss to ensure that the latent feature distribution is similar to the Gaussian prior $p(h) = \mathcal{N}(0, 1)$. The weight of $L_1$ loss in triplane and depth is 1, the weight of LPIPS loss in the image is 1, and the weight of KL loss is $1 \times 10^{-5}$. We randomly sample camera poses during rendering, with the sampling ranges set to pitch in $[-0.25, 0.65]$ radians, yaw in $[-0.78, 0.78]$ radians, and roll in $[-0.25, 0.25]$ radians. The visual results of our VAE are shown in Figure~\ref{fig:sup_vae}.

\subsection{DiT}
The VAE compresses the triplane into $z_t \in \mathbb{R}^{64 \times 64 \times 4 \times 8}$. The DiT reshapes $z_t$ to $64 \times 256 \times 8$, adds positional embeddings, and then flattens it before feeding it into the Transformer for training. Following the approach in Direct3D~\cite{wu2024direct3d}, at each DiT block, we concatenate DINO tokens with the flattened $z_t$ and pass them through a self-attention mechanism to capture the intrinsic relationships between the DINO tokens and $z_t$. Afterward, we discard the image tokens, retaining only the noisy tokens for input to the next module.  Moreover, to reduce the number of parameters and computational cost, we adopt adaLN-single, as introduced in PixArt~\cite{chen2023pixartalpha}. This method predicts a set of global shift and scale parameters $P = [\gamma_1, \beta_1, \alpha_1, \gamma_2, \beta_2, \alpha_2]$ using time embeddings. A trainable embedding is then added to $P$ in each block for further adjustment. During training, the batch size is set to 1536,  and the training is conducted over 48 Tesla A100 GPUs (batch size 32 for each GPU), each with 80GB of memory, for a total of 5 days.

\subsection{Motion-Aware Cross-Domain Renderer}
\begin{figure}[t]
\begin{center}
\includegraphics[width=0.96\linewidth]{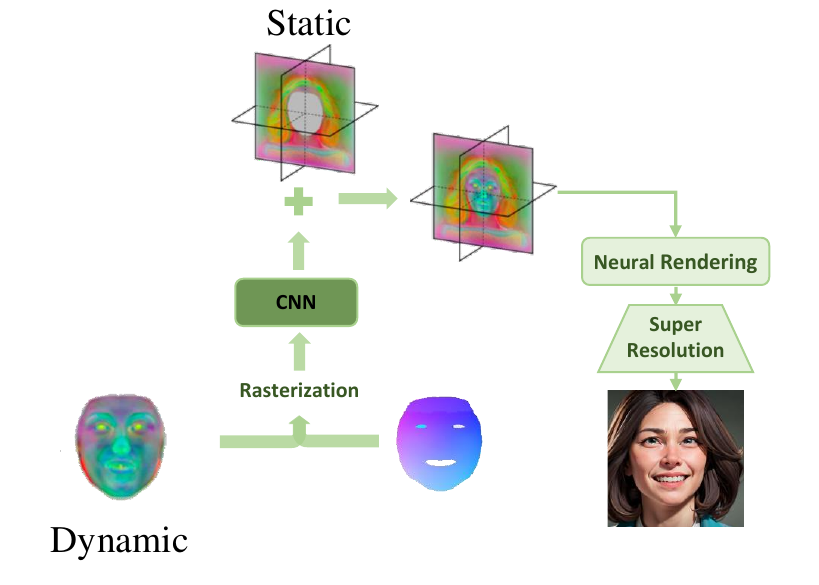}
\caption{Visualization of rendering process of Next3D. After rasterization, a CNN is employed to remove artifacts introduced during the rasterization process, which is critical for final performance, as mentioned in the Next3D~\cite{sun2023next3d}.  }
\vspace{-0.17in}
\label{fig:sup_next3D}
\end{center}
\end{figure}

During the Next3D rendering process in Figure.~\ref{fig:sup_next3D}, a CNN is used to refine the dynamic components after rasterization, eliminating artifacts introduced in the rasterization stage (e.g., teeth completion, identity leakage). When training Next3D for different domains, we fine-tune this CNN, as well as the MLPs used in both super-resolution and neural rendering. Therefore, a unified renderer is required to handle parametric triplanes from various domains and mitigate issues caused by rasterization. 

As mentioned in our main paper, we find a simple CNN can not handle the cross-domain parametric triplanes, and we propose the motion-aware cross-domain renderer. To train the motion-aware cross-domain renderer, we use the trained 4DGAN to generate the 4D images (i.e., multi-view, multi-expression images of the same individual), and we are able to simultaneously obtain the corresponding depth, parametric triplane, and rendering features. The data is separated into static and dynamic parts similar to Portrait4D~\cite{deng2024portrait4d}, as mentioned in our main paper. The overall training objective of our renderer is defined as follows:
\begin{equation}
\mathcal{L} = \mathcal{L}_{\text{re}} + \mathcal{L}_{\text{f}} + \mathcal{L}_{\text{tri}} + \mathcal{L}_{\text{depth}} + \mathcal{L}_{\text{opa}} + \mathcal{L}_{\text{adv}},
\label{eq:training}
\end{equation}
where $\mathcal{L}_{\text{re}}$ represents a combination of the LPIPS and $L_1$ distances between the generated image $I_{o}$ and its corresponding ground truth. $\mathcal{L}_{\text{tri}}$ measures the $L_1$ difference between the generated triplane features and their ground truth. $\mathcal{L}_{\text{f}}$ computes the $L_1$ difference between the generated rendering features and their respective ground truth. $\mathcal{L}_{\text{depth}}$ evaluates the $L_1$ difference between the generated depth map and its ground truth counterpart. $\mathcal{L}_{\text{opa}}$ is the $L_1$ difference between the predicted opacity and the ground truth. Finally, $\mathcal{L}_{\text{adv}}$ represents the adversarial loss between $I_{o}$ and the ground truth image, utilizing the discriminator from the Next3D model.

The loss balancing weights for each term in Eq.~(\ref{eq:training}) are set to 1, 1, 0.1, 1, 1, and 0.01 for $\mathcal{L}_{\text{re}}$, $\mathcal{L}_{\text{f}}$, $\mathcal{L}_{\text{tri}}$, $\mathcal{L}_{\text{depth}}$, $\mathcal{L}_{\text{opa}}$, and $\mathcal{L}_{\text{adv}}$, respectively. For the first 1000K images, $\mathcal{L}_{\text{adv}}$ is not applied, and the parameters in both the neural renderer and super-resolution components are kept fixed. After 1000K images, $\mathcal{L}_{\text{adv}}$ is introduced, and the trainable parameters of the neural renderer and super-resolution modules are unfrozen. We employ volume rendering with 48 coarse samples and 48 fine samples per ray. The initial volume rendering resolution is set to $64^2$ for the first 1000K images, gradually increasing to $128^2$ as training progresses. The model is trained on a total of 8 million images. We utilize the Adam optimizer with $(\beta_1, \beta_2) = (0.9, 0.999)$ and a learning rate of $1 \times 10^{-4}$ across all networks. The batch size is set to 96, with an even split between dynamic and static data. The training is conducted over 24 Tesla A100 GPUs, each with 80GB of memory, for a total of 4 days.

\section{Additional Comparisons and Visual Results}\label{appendix:results}
\begin{table}[t]
\centering
\resizebox{1\linewidth}{!}{
\begin{tabular}{@{}lcccc@{}}
\toprule
\textbf{Model} & \multicolumn{4}{c}{\textbf{Trained Domains / Untrained Domains}} \\ \cmidrule(lr){2-5} 
 & \textbf{Sharpness} & \textbf{Temporal} & \textbf{Expression} & \textbf{Identity } \\ \midrule
LivePortrait & 3.625 / 2.5 & 3.625 / 2.5 & 3.5 / 1.5 & 3.5 / 1.5 \\
Xportrait & 2.375 / 1 & 1.625 / 1 & 2 / 1 & 1.875 / 1.5 \\
Invertavatar & 2.625 / 2.5 & 2.25 / 2.5 & 2.375 / 2 & 2.75 / 2.5 \\ 
Portrait4D & 1.875 / 3.5 & 2.5 / 3 & 2.125 / 2.5 & 2.375 / 2 \\ \midrule
Ours & \textbf{4.625} / \textbf{5}& \textbf{4.25}  / \textbf{5} & \textbf{4.375} / \textbf{4.5} & \textbf{4.125} / \textbf{5} \\ \bottomrule
\end{tabular}%
}
\caption{User Study.}
\label{tab:userstudy}

\end{table}

\subsection{User Study}
For a more comprehensive evaluation, we conducted a user study with 10 participants, who were asked to assess image sharpness, temporal consistency, expression consistency, and identity consistency. They did so by selecting the best method while reviewing 12 cross-ID reenactment results generated by different approaches.

For each evaluation criterion, participants were presented with five videos, each corresponding to the results produced by a different method. They were instructed to rate the videos on a scale from 1 to 5, where 5 indicates the highest quality and 1 the lowest. Multiple methods could receive the same score. As shown in Table~\ref{tab:userstudy}, our method exhibits significant advantages over the others. 
\subsection{Visual Comparisons}
\begin{figure*}[t]
\begin{center}
\includegraphics[width=0.96\linewidth]{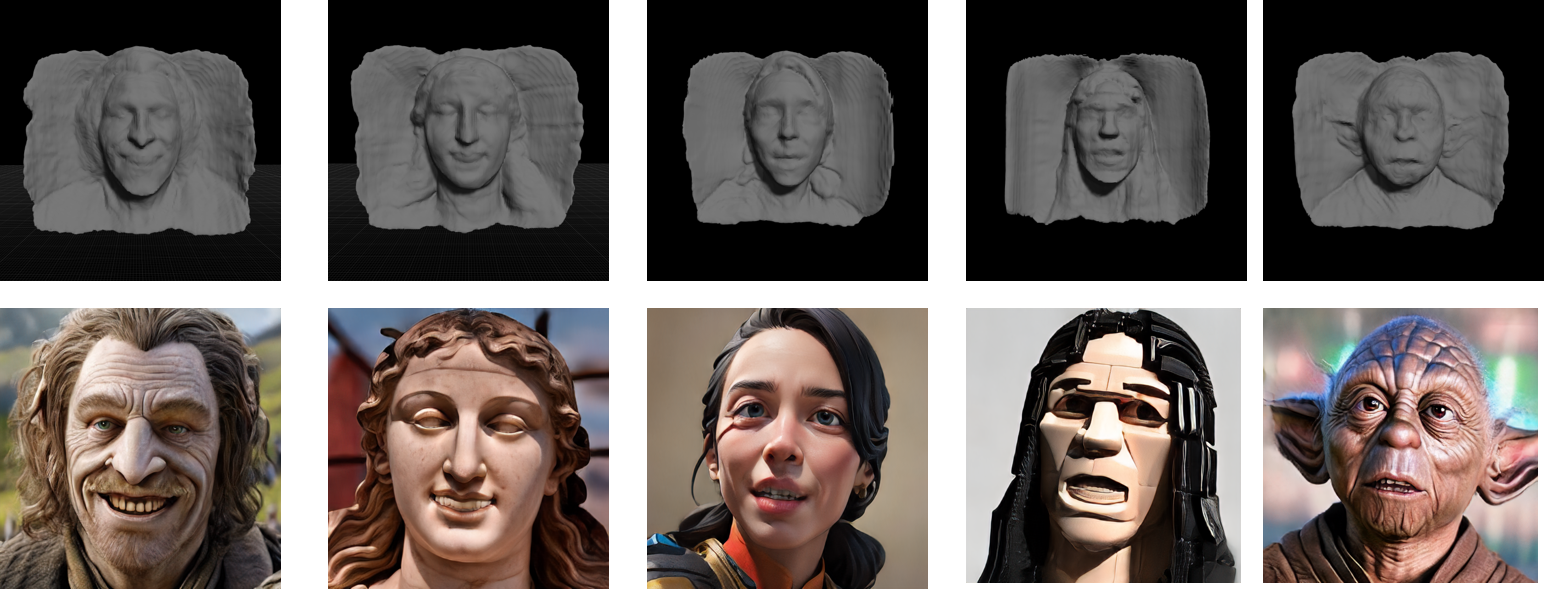}
\caption{The geometry results of our method.}
\vspace{-0.17in}
\label{fig:geometry}
\end{center}
\end{figure*}

\begin{figure*}[t]
\begin{center}
\includegraphics[width=0.96\linewidth]{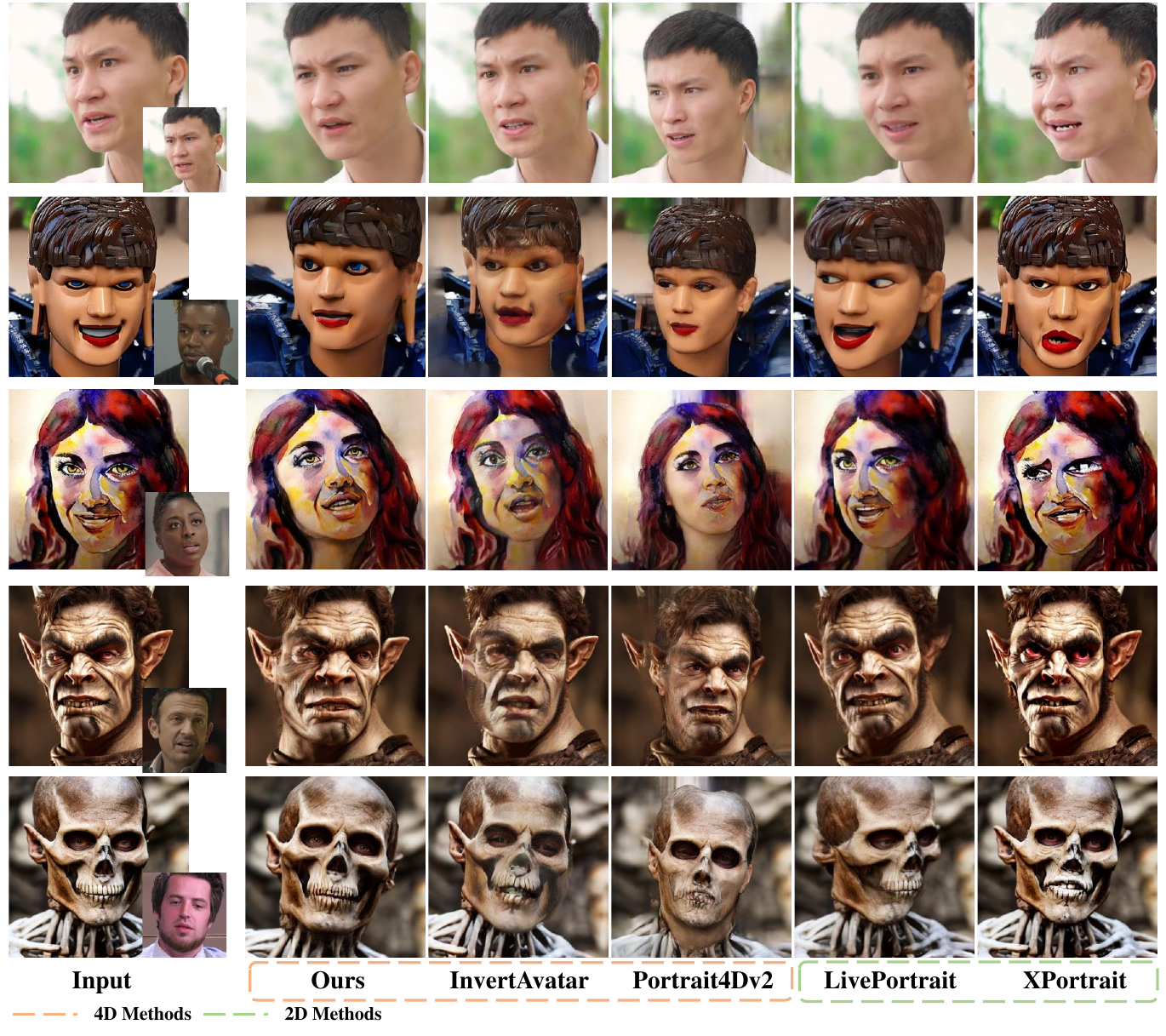}
\caption{Qualitative comparison with state-of-the-art methods. The leftmost column of the figure presents the input images, with the bottom-right corner indicating the target image. The first row illustrates the results of self-reenactment, while the subsequent rows showcase the results of cross-reenactment. Our method demonstrates superior performance in terms of expression and pose consistency, as well as identity preservation. For more visual results, please refer to our \textcolor{orange}{video results}. }
\vspace{-0.17in}
\label{fig:sup_compare}
\end{center}
\end{figure*}

In Figure~\ref{fig:sup_compare}, we present additional visual comparisons, demonstrating that our method achieves superior performance. Moreover, we present our geometric results in Figure~\ref{fig:geometry}. For more visual results, please refer to our \textcolor{orange}{video results}.

\end{document}